# To Go or Not To Go?
# A Near Unsupervised Learning Approach For Robot Navigation

Noriaki Hirose[1][*], Amir Sadeghian[1][*], Patrick Goebel[1] and Silvio Savarese[1]

*Abstract*—It is important for robots to be able to decide whether they can go through a space or not, as they navigate through a dynamic environment. This capability can help them avoid injury or serious damage, e.g., as a result of running into people and obstacles, getting stuck, or falling off an edge. To this end, we propose an unsupervised and a near-unsupervised method based on Generative Adversarial Networks (GAN) to classify scenarios as traversable or not based on visual data. Our method is inspired by the recent success of data-driven approaches on computer vision problems and anomaly detection, and reduces the need for vast amounts of negative examples at training time. Collecting negative data indicating that a robot should not go through a space is typically hard and dangerous because of collisions; whereas collecting positive data can be automated and done safely based on the robot's own traveling experience. We verify the generality and effectiveness of the proposed approach on a test dataset collected in a previously unseen environment with a mobile robot. Furthermore, we show that our method can be used to build costmaps (we call as "GoNoGo" costmaps) for robot path planning using visual data only.

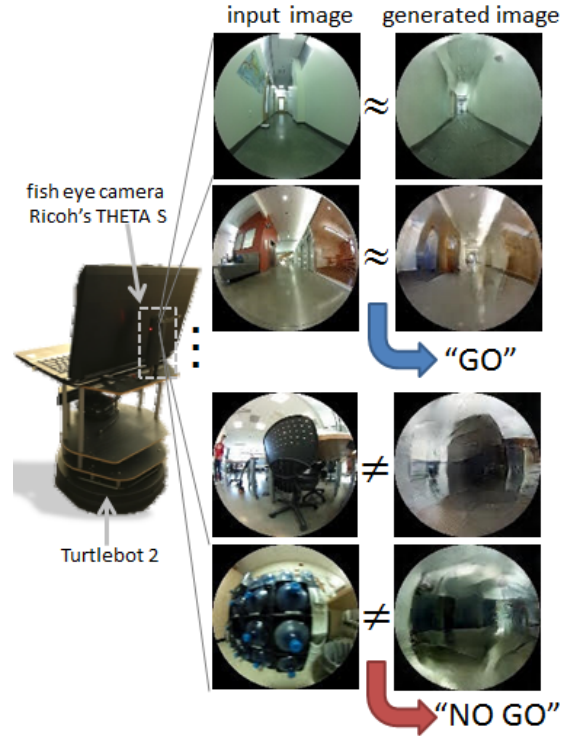

Fig. 1. Basic idea of our proposed approach.

## I. INTRODUCTION

Robot navigation is essential for many tasks, such as guiding people through a space [1], carrying heavy luggage for shopping [2], [3], [4], automated delivery, or environmental inspection. To successfully navigate in many of these circumstances, robots need to adapt to the presence of dynamic obstacles and changes in their environment.

Motivated by the recent success of neural network models in control and perception applications, this work explores the use of deep learning for mobile robot navigation with a single camera. In particular, we study the problem of specifying if a wheeled robot can go through a space or not. We further refer to these cases as *GO* or *NO GO* situations, respectively. Making the right decision in this problem can prevent robots from colliding with objects, injuring people, getting stuck in constrained spaces, or falling over an edge.

Building accurate neural network models often require a large amount of annotated data, but collecting this data can be both time-consuming and costly. In applications such as robot navigation, incorrect annotations made by human error can further cause serious damage to both the robot and its environment. Besides, collecting negative examples of situations in which a robot should not traverse a space can be challenging and dangerous. Some of these spaces are impossible for a robot to go through. Others may involve expensive collisions and injuries. The alternative idea is to use sensors like a bumper for collecting negative data. However, bumpers may not work well with small obstacles nor prevent the robot from falling down an edge. IR sensors can also be used as "cliff sensors, but the consequences of false detections are often too costly to depend on them.

The main insight of our work is that we can build a reliable deep model for the GO or NO GO problem using vision and an unbalanced dataset of examples. This dataset is mainly composed of positive examples that can be automatically collected safely based on the robot's traveling experience, e.g., under human supervision. Moreover, we build this dataset using an onboard and off-the-shelf fisheye camera. This type of sensor makes our solution practical and cheap for mobile platforms in comparison to using more expensive sensors, such as LIDARs.

Inspired by prior work on anomaly detection [5], we propose two methods based on Generative Adversarial Networks (GAN) to classify the scenarios which a robot can go through or not [6], [7]. One method is unsupervised; the other, which is an extension, is near unsupervised.

The proposed unsupervised method uses GAN to train a generator function ($Gen$) that generates images in the manifold of the positive dataset. The difference between the real image $X$ and the generated image $X'$ through the trained generator is used to classify whether the real image is positive or not. Note that, the generator trained on the positive images can not generate negative images. Hence, it is expected that the difference between the real image and the generated image ($|X - X'|$) for the negative examples to be bigger than the positive examples. However, the two main limitations for implementing this approach on a real-time mobile robot are high computational time and low accuracy.

* indicates equal contribution
[1]N. Hirose, A. Sadeghian, and S. Savarese are with Stanford AI Lab., Computer Science, Stanford University, 353 Serra Mall, Gates Building, Stanford, CA, USA `hirose@cs.stanford.edu`

To solve the first limitation, we design another network as the inverse generator ($Gen^{-1}$) to produce the generated image ($X'$) corresponding to the input image ($X$) in real time [8]. To address the second limitation, we propose a near unsupervised method. And use a relatively small amount of negative and positive annotated data. The annotated dataset is less than 1 percent of the whole positive dataset used for our unsupervised learning method. This small amount of data annotation is acceptable for our problem and can help to improve the performance significantly. All the components in our method are performed in a feed-forward manner and are applied in real time with high accuracy, applicable on real-time mobile robots.

In this work, we use a fisheye RGB camera image as the input ($X$) for our method. Fisheye cameras can efficiently capture every angle of the surrounding environment. It enables the robot to see up, down, and side areas using a single camera. Moreover, the cost of a Fisheye camera is much less than a 3D LIDAR.

The rest of the paper is organized as follows. Section II first describes related work on deciding whether robot's location is traversable or not. The proposed data-driven approach and its evaluation are then presented in Sections III and IV. The latter section we show a detailed experimental study of our method and introduce a novel costmap called GoNoGo generated by our method for the task of path-planning. Section V finally concludes this paper and discusses future work.

## II. RELATED WORK

Prior work has proposed different approaches to classify GO or NO GO situations and perform obstacle avoidance. We briefly describe these works and their connections to our method.

### A. Measurement Methods

Different sensors have been used in the past to estimate whether a physical space is traversable or not. For example, Suger et al [9] used a 3D LIDAR to design a grid map for robot navigation. This map was built using a naive Bayes classifier. Pfeiffer et al. [10] proposed an imitation learning method to learn how a robot should navigate a space based on expert demonstrations and 2D LIDAR data. Borenstein et al. [11] rather proposed a histogram method to avoid obstacles using an ultrasonic sensor, and Er et al. [12] trained a neuro-fuzzy controller to mimic innate behavior using IR sensors.

Other prior methods have relied on image data for obstacle detection and avoidance. For instance, Ulrich et al. [13] used a monocular camera and color differences to detect obstacles on the floor. Other efforts have used monocular cameras to estimate depth [14], [15], [16], which can then be used to avoid obstacles.

Different to these lines of work, we propose to use a single RGB fisheye camera for the GO or NO GO problem. This type of camera can capture every angle of the surrounding environment and is significantly cheaper than a LIDAR. Furthermore, our results suggest that depth information is not necessarily needed for this problem.

### B. Deep Learning Techniques

Nowadays, deep learning techniques have been successfully extended to many fields such as robotics, computer vision [17], modeling [18], [19], [20], control [21], voice recognition and so on. We can divide previous works on the obstacle avoidance into two categories, 1) imitation of human behavior, and 2) Near unsupervised and supervised learning.

*1) imitation of human behavior:* LeCun et al [22] train neural networks to mimic human behavior for autonomous vehicles. The steering input of the human drivers is collected for their supervised learning techniques. S. Ross et al [23] trains DAgger [24] to imitate a human being's behavior for obstacle avoidance using a drone. Tai et al [25] uses neural networks to mimic the traveling velocity and turning angular velocity from the joypad control of a wheeled robot. Huang et al [26] proposes an approach to solve the problem of autonomous mobile robot obstacle avoidance using reinforcement learning neural networks. Giusti et al [27] trains a network to decide go straight, turn right or turn left for a drone. The camera images with annotation are collected by a trail in a forest. However, referenced human motion in these methods often includes not appropriate behavior for the training, which may cause an accident. As opposed, we only collect a positive dataset that is automatically annotated by the robot's own experience (without any errors) and do a classification task. Therefore, our method can suppress the possibility to learn the wrong behavior and reduce the risk of an accident significantly.

*2) Near Unsupervised and Supervised Learning:* Gandhi et al [28] collects negative images ("NO GO") by crashing drones into obstacles and uses a neural network to classify scenes into GO or NO GO. However, in our problem there are many situations that the wheeled robot can not have a crash, e.g. falling from an edge. Elkan et al [29] proposes the PU learning method to distinguish between positive and negative samples only using a positive and unlabeled dataset. Schlegl et al [5] applies an unsupervised learning approach based on GAN for anomaly detection. These approaches [5], [29] don't need to have the annotation process for the negative dataset. However, it is difficult to apply current PU learning methods[29] to our problem because they work well in the scenarios where the distribution or the domain of positive and negative examples are limited and simple. On the other hand, the calculation speed and the accuracy of most methods such as [5] are not enough for the real-time mobile robot. We evaluate the performance of this baseline in more details in the later section.

To address the performance and computation costs of the baseline methods, we propose an unsupervised and a near unsupervised learning method. We also avoid the time-consuming and costly annotation process.

## III. LEARNING TO "GO" OR "NO GO"

### A. Overall Architecture

Figure 2 shows the overall architecture of our proposed approach. First, our approach tries to generate an image $X'$,

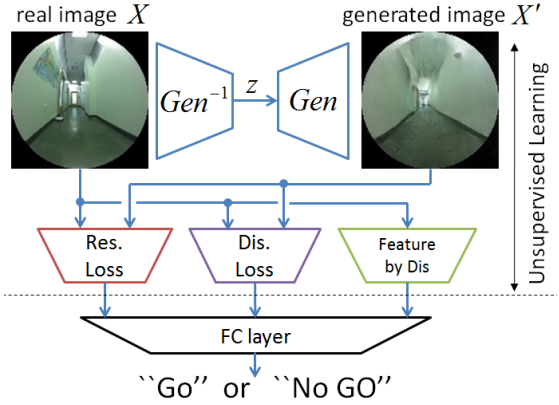

Fig. 2. Overview of our proposed approach. $X$ is the real input image, $X'$ is generated image, $Gen$ is the generator trained by GAN, $Gen^{-1}$ is trained inverse generator to realize $X = X'$, $Dis$ is the discriminator trained by GAN, FC Layer is a fully connected layer trained by a small amount of annotated data.

TABLE I
PARAMETERS OF GENERATOR $Gen$.

|        | filter size  | stride | output size              | function |
|--------|--------------|--------|--------------------------|----------|
| Input  | –            | –      | 100                      | –        |
| FC1    | –            | –      | $8 \times 8 \times 512$  | Linear   |
| Dconv2 | $4 \times 4$ | 2      | $16 \times 16 \times 256$| Relu     |
| Dconv3 | $4 \times 4$ | 2      | $32 \times 32 \times 128$| Relu     |
| Dconv4 | $4 \times 4$ | 2      | $64 \times 64 \times 64$ | Relu     |
| Dconv5 | $4 \times 4$ | 2      | $128 \times 128 \times 3$| Relu     |

TABLE II
PARAMETERS OF DISCRIMINATOR $Dis$.

|        | filter size  | stride | output size               | function |
|--------|--------------|--------|---------------------------|----------|
| Input  | –            | –      | $128 \times 128 \times 3$ | –        |
| Conv1  | $4 \times 4$ | 2      | $64 \times 64 \times 64$  | Elu[30]  |
| Conv2  | $4 \times 4$ | 2      | $32 \times 32 \times 128$ | Elu      |
| Conv3  | $4 \times 4$ | 2      | $16 \times 16 \times 256$ | Elu      |
| Conv4  | $4 \times 4$ | 2      | $8 \times 8 \times 512$   | Elu      |
| FC5    | –            | –      | 2                         | softmax  |

which corresponds to the real input image $X$ through the manifold of the positive dataset. The generator function $Gen$ is trained by a GAN and outputs images in the manifold of the positive examples. $Gen^{-1}$ is the inverse generator to search for the appropriate latent vector $z$ to express $X$. We apply $Gen^{-1}$ to decrease the computational load instead of the iterative back-propagation method used in previous base line methods [5].

We then extract the following three features from $X$ and $X'$ to classify the scene observed in the input image as GO or NO GO:

(R) **Residual Loss:** $||X - X'||$,
(D) **Discriminator Loss:** $||f(X) - f(X')||$,
(F) **Feature by Discriminator:** $f(X)$,

where $f$ is the last convolution layer features of our GAN's discriminator. Because $Gen$ and $Gen^{-1}$ are trained only on positive data samples [5], our method expects that the residual loss "R" and the discriminator loss "D" are large when the input image is a negative example. However, for some negative examples, R and D are not discriminative enough for accurate "GO", "NO GO" classification. Thus, we modify the weight of salient areas as shown in section III-D. To improve the performance of our method furthermore, we train an FC layer with a small amount of annotated data.

### B. Manifold of "GO" Image

Figure 3 depicts our GAN, which is constructed by two adversarial modules, a generator, $Gen$ and a discriminator, $Dis$. Here, $z$ is the noise generated by a normal distribution. $Dis$ is trained to decide whether the input is real or generated image. On the other hand, $Gen$ is trained to fool $Dis$. $Dis$ and $Gen$ are simultaneously trained by the following a two player min-max game:

$$\min_{Gen} \max_{Dis} V(Dis, Gen) = E_{X \sim p_{data}(X)}[\log Dis(X)]$$
$$+ E_{z \sim p_z(z)}[\log(1 - Dis(Gen(z)))], \quad (1)$$

where $p_{data}$ and $p_z$ are the distribution of $X$ and $z$, respectively. The training dataset $X$ is composed of images with positive labels only.

In the proposed approach, $Gen$ and $Dis$ are designed as standard Convolutional Neural Networks (CNN) listed in Table I and II. Here, "FC" is a fully connected Layer, "Dconv" is a de-convolutional layer, and "Conv" is a convolutional layer. Batch normalization is applied after each convolutional and deconvolutional layer.

### C. Training of the Inverse Generator

To generate an image similar to the input data, an adequate value for the noise $z$ has to be found. Prior work by Schlegl et al. [5] applies an iterative back-propagation procedure for 500 times to minimize the following cost function under the fixed $Gen$ and $f$:

$$\mathcal{L}(z) = (1 - \lambda) \cdot \mathcal{L}_R(z) + \lambda \cdot \mathcal{L}_D(z), \quad (2)$$

where the residual loss $\mathcal{L}_R(z)$ and the discriminator loss $\mathcal{L}_D(z)$ are defined as follows:

$$\mathcal{L}_R(z) = ||X - Gen(z)||, \quad (3)$$
$$\mathcal{L}_D(z) = ||f(X) - f(Gen(z))||. \quad (4)$$

The parameter $\lambda$ in eq. (2) is a weighting factor for $\mathcal{L}_R(z)$ and $\mathcal{L}_D(z)$. Unfortunately, the computational load of the iterative back-propagation procedure used to minimize $\mathcal{L}(z)$ is too expensive and it's not applicable on a real-time mobile robot.

In order to speed up this process, we train and apply the inverse generator $Gen^{-1}$ to find the appropriate noise $z$, as shown in Fig. 4 [8]. The structure of the network corresponding to $Gen^{-1}$ is listed in Table III. This is the same design as $Dis$ except for the last FC5. The output size of FC5 in our $Gen^{-1}$ is set as 100, in order to match the size of $z$. $Gen^{-1}$ is trained only on positive data by minimizing the cost function $\mathcal{L}(z)$ under a fixed $Gen$..

### D. Weighting for Unsupervised Learning

The base line method uses the following score $A(x)$ for classification:

$$A(X) = (1 - \lambda) \cdot R_s(X) + \lambda \cdot D_s(X), \quad (5)$$

TABLE III
PARAMETERS OF INVERSE GENERATOR $Gen^{-1}$.

| | filter size | stride | output size | function |
|---|---|---|---|---|
| Input | – | – | $128 \times 128 \times 3$ | – |
| Conv1 | $4 \times 4$ | 2 | $64 \times 64 \times 64$ | Elu[30] |
| Conv2 | $4 \times 4$ | 2 | $32 \times 32 \times 128$ | Elu |
| Conv3 | $4 \times 4$ | 2 | $16 \times 16 \times 256$ | Elu |
| Conv4 | $4 \times 4$ | 2 | $8 \times 8 \times 512$ | Elu |
| FC5 | – | – | 100 | linear |

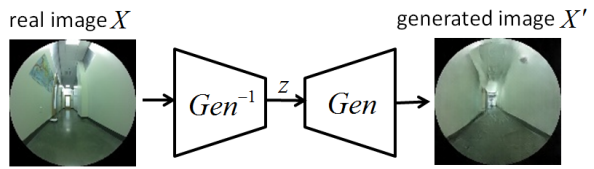

Fig. 4. Neural Network Structure for training $Gen^{-1}$

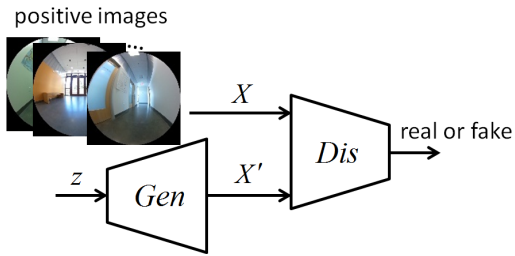

Fig. 3. Neural Network Structure of DCGAN

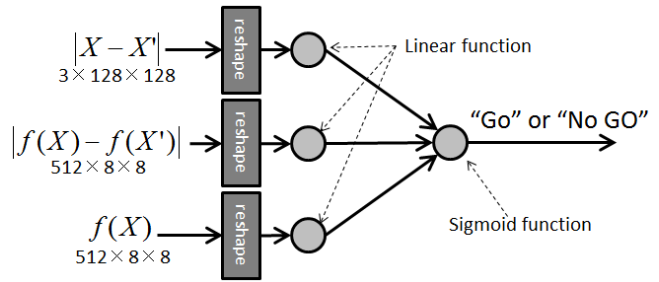

Fig. 5. Neural Network structure of "FC layer"s

where the residual score $R_s(X)$ and the discriminator score $D_s(X)$ are defined as $\mathcal{L}_R(z_\Gamma)$ and $\mathcal{L}_D(z_\Gamma)$, respectively. $z_\Gamma$ is the latent space vector representing the input image. The base line method classifies the scene as GO or NO GO by thresholding $A(X)$ as follow:

$$t_d = \begin{cases} 1 & \text{"GO"} \quad (A(x) < a_{th}) \\ 0 & \text{"NO GO"} \quad (A(x) \geq a_{th}) \end{cases}, \quad (6)$$

where $t_d$ is decision flag for GO (=1) or NO GO (=0). The threshold value $a_{th}$ is set to 0.17. The base line method can not precisely distinguish between positive and negative images. To address this problem, we modify $R_s(X)$ and $D_s(X)$. Basic idea is weighting the salient areas of the image more instead of simple L2 norm of the difference in (3) and (4) as follows:

$$\mathcal{L}_R(z) = ||W_R \circ (X - Gen(z))||, \quad (7)$$
$$\mathcal{L}_D(z) = ||W_D \circ (f(X) - f(Gen(z)))||, \quad (8)$$

where $W_R \epsilon \mathbb{R}^{3 \times 128 \times 128}$ and $W_D \epsilon \mathbb{R}^{512 \times 8 \times 8}$ are the weighting matrices, and $\circ$ indicates pointwise product function. Using cross validation we found that the salient area for classifying GO or NO GO is the bottom area on the image, which corresponds to the close area on the floor in front of the robot. Thus, we gave more weight to one eighth bottom area of the image.

*E. FC Layer*

The accuracy of unsupervised learning method reported in Table IV is not high enough to be implemented on an actual robot. Therefore, we trained an FC layer shown in Fig. 5 with a tiny amount of positive and negative annotated data. The gray circles in Fig. 5 corresponds to the linear layers.

The tensor inputs $|X - X'| \epsilon \mathbb{R}^{3 \times 128 \times 128}$, $|f(X) - f(X')| \epsilon \mathbb{R}^{512 \times 8 \times 8}$, and $f(X) \epsilon \mathbb{R}^{512 \times 8 \times 8}$ are reshaped into vectors to calculate the product of the weight and the sum of bias at each node. And the output of these layers are scalars. Therefore, a $3 \times 1$ feature vector is given into the last layer shown as the grey circle at the right side in Fig.5. The sigmoid function is given as the activation function to estimate the probability GO or NO GO at the last node. The reason why the FC layer has a simple network structure is to avoid over-fitting, because we are only using a small amount of annotated data. In addition to our simple network structure, we stop training early for the FC layer to avoid over-fitting.

IV. EXPERIMENTAL RESULTS

*A. Robot Platform*

Left side of Fig.1 depicts the robot used for our experiments. This is a "Turtlebot 2" platform [31] with a "THETA S" fish eye camera by Ricoh [32]. Turtlebot 2 is a common research platform that is easy to use with the Robot Operating System (ROS) [33].

Ricoh's THETA S can efficiently capture every angle of the surrounding environment around the robot with a full HD resolution ($3 \times 1920 \times 1080$) at 15 fps. It enables the robot to have full visibility of the area above, below and around it with a single camera. This wide view is very important to capture the environment and decide to GO or NO GO.

*B. Data Collection*

Figure 6 shows the map of the engineering quad at Stanford University. Here, the red rectangles indicate the buildings where data was collected for the training set, and the blue rectangles indicate those where the test set was gathered. The length of the video collected per location for training and evaluating the proposed approach is shown beside the highlighted rectangles.

In each building, we controlled the robot using a gamepad and collected videos at 3 fps. Although the THETA S has 2 fisheye cameras, one in front and one in the back, we only used the front camera. The total duration of the data collected for the experiments was about 7.2 hours.

The recordings led to a dataset of 78,711 useful images for the present evaluation. The images were cropped and resized

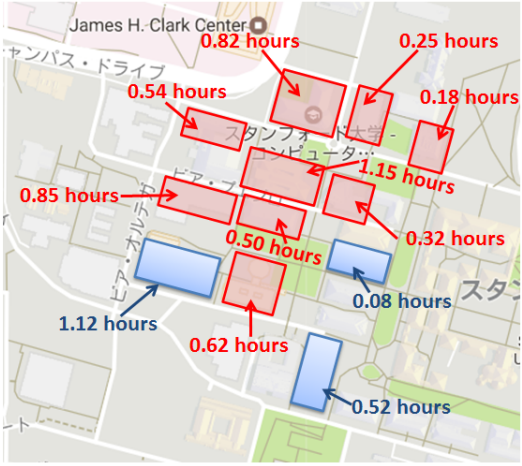

Fig. 6. Amount of data captured per location for our experiment. The information was overlaid on a view of the campus from Google Maps [34].

---
**Algorithm 1** Automatic labeling of positive dataset
---
1: **for** $i = 0$ to $N$ **do**
2:    **if** $v(x) > v_{th}, \forall x \in [i-p, \cdots, i, \cdots i+f]$ **then**
3:       label of $X(i)$ is positive
4:    **else**
5:       label of $X(i)$ is not defined (unlabel)
6:    **end if**
7: **end for**
---

to a resolution of $3 \times 128 \times 128$ pixels for our experiments. The images were also flipped horizontally to augment our dataset.

### C. Annotation

We used the robot's velocity to automatically identify situations in which the robot could traverse the space in front of it. More specifically, the images that were captured within a time window in which the robot's velocity $v$ was bigger than a threshold value $v_{th}$ were automatically labeled as positive examples in our dataset. This procedure is detailed in Algorithm 1. The hyper parameters $v_{th}$, $p$ and $f$ were set to 0.3 m/s, 5, and 3, respectively. This resulted in 53598 and 17968 positive images for training and testing, respectively.

In addition, two small datasets of positive and negative annotated images are required, one for training the FC layer and other for evaluation of our overall method. For the positive dataset, we randomly select 400 images from the annotated images using Algorithm 1. The negative annotations are given by hand. A candidate set of negative images were randomly chosen from the unlabeled dataset. And 400 images from these candidates were hand labeled by the authors. The above process was done both for training and test data. The amount of hand-labeled data for training the FC layer (near supervised method) was less than 1% of the overall positive training dataset used for training our GAN (unsupervised learning method).

### D. Training

There are three steps for training our method. Two of them only use positive data (unsupervised learning) i) GAN in Fig.3 and ii) inverse generator $Gen^{-1}$ in Fig.4. The third step is part of the supervised learning method which is training the FC layer in Fig.5 to improve the accuracy of our near unsupervised learning method. We implemented and trained all our methods using "chainer" deep learning framework [35]. For training, we use a mini-batch size of 100 and the optimization method is ADAM[36].

### E. Visualization

*1) Performance of the trained $Gen$ and $Gen^{-1}$ models:*
In order to verify the performance of trained $Gen$ and $Gen^{-1}$, Fig.7 shows a set of real images ($X$) in [a], and the corresponding generated images ($X'$) in [b] for the positive images. Similarly, Fig.8 shows the real images ($X$) and the corresponding generated images for the negative images. The images in Fig.7 and 8 are randomly chosen from the test set.

As we can see in Fig.7[a] and [b], the generated images $X'$ are similar to their corresponding real images $X$ for the positive dataset, except some small differences. For example, a person in the upper center image, a windows frame in the middle left image, and a small white box in the bottom center image in $X$ are partially removed in the generated images $X'$. However, the general look and the color of the images is almost same.

On the other hand, Fig.8 depicts the big difference between $X$ and $X'$ for the negative dataset. The generated images in Fig.8[b] look like corridors or hallways, which are the typical in the positive dataset, although the inputs were negative images. The obstacles, like trash box, wall, fence, stairs, wooden furniture, and chair are disappeared in the generated images $X'$, because $Gen$ is the manifold of the positive images and can not generate NO GO situations.

Note that in some cases, the difference between the negative image that was input to our system and the one it generated is not very large. For example, in the scenario depicted in the bottom left corner of Fig. 8, some of the stairs get washed out in the generated image $X'$. However, the overall look of the input and the generated images is similar. Or, in the bottom right image, the brown box in $X$ is turned into a brown corridor in $X'$. Thus, relying solely on an unsupervised learning method to distinguish between positive and negative examples just by comparing $X$ and $X'$ can be hard. As we show in our ablation study in the next section, the final FC layer of our model can handle these situations robustly.

*2) Saliency map:* We visualize the behavior of our overall trained neural network using a saliency map[37]. Figure 9[a] is the mean of all the saliency maps for the positive class. Almost all the white area (most salient) is shown inside of the red lines. The center and upper area of the image are not that important for predicting the GO or NO GO classes. In particular, the upper area of the input images was typically the ceiling of the room where the robot navigated through, while the center area was typically a space farther away from the robot. In contrast, the bottom area was often occupied by the floor in front of the robot. The right and left sides were often a wall of the room or the corridor in the vicinity of the robot. These areas are the most important part of the image to look.

We also visualized the weights for the residual loss of the FC layer in Figure 9[b]. Four corners and center area is almost black, while inside of two red curves is white, which has bigger weight to predict GO or NO GO. This result indicates that the residual loss R gives a positive effect to the whole system of our proposed approach.

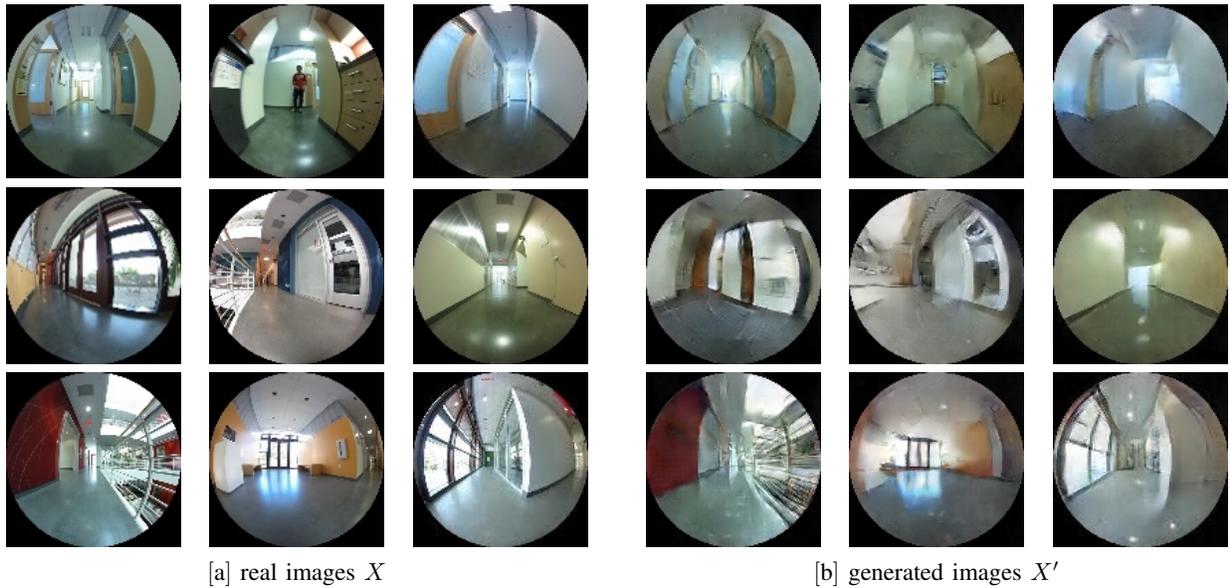

[a] real images $X$                [b] generated images $X'$

Fig. 7. Real images $X$ and generated images $X'$ for test set with positive label.

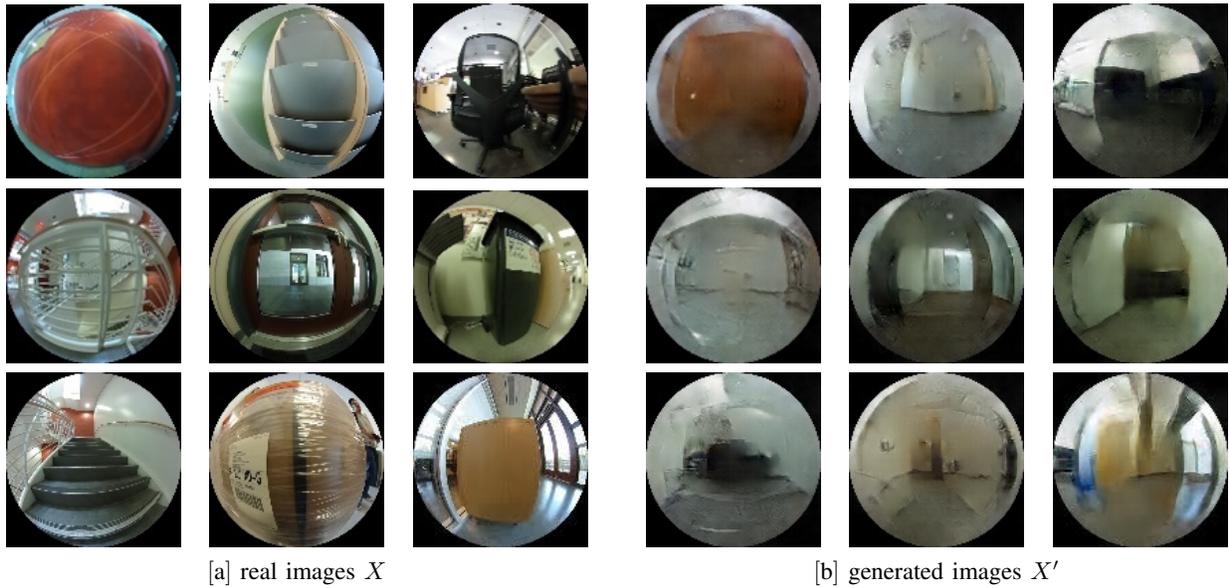

[a] real images $X$                [b] generated images $X'$

Fig. 8. Real images $X$ and generated images $X'$ for test set with negative label.

*F. Ablation Study*

Table IV shows the results from our evaluation on the test set using different sets of components: (R) Residual loss, (D) Discriminator loss, and (F) Features by the discriminator. The table also shows results for unsupervised [5] and supervised [38] baseline methods.

We use accuracy, recall, precision, and f1 score as metrics to evaluate our performance. Also, the frequency, which corresponds to the calculation speed, and memory size are also listed to understand whether these techniques are applicable on a real mobile robot. For the measurement of frequency and memory size, we ran our experiments on a Geforce GTX TITAN X GPU. We further describe the baselines used in our evaluation and discuss the results presented Table IV in the following sections.

*1) Unsupervised learning:* We use [5] as an unsupervised baseline method. Our method for the unsupervised learning is using inverse generator $Gen^{-1}$ to search the appropriate noise $z$ and to apply weighting modifications shown in the section III-D.

As shown in Table IV, our method performs better than the unsupervised baseline method. Moreover, the computation speed is improved by the inverse generator $Gen^{-1}$. However, an accuracy of 72.5 % with unsupervised learning is low for practical purposes. The problem is that sometimes the difference between $X$ and $X'$ is not informative enough for classification purposes. For positive input images, the generator sometimes washes out small, but important details. For some negative input images, the difference between $X$ and $X'$ can be small, as discussed in the previous section.

TABLE IV
ANALYSIS OF OUR METHOD ON THE TEST SET USING DIFFERENT SET OF COMPONENTS (R) RESIDUAL LOSS, (D) DISCRIMINATOR LOSS, AND (F) FEATURE OF DISCRIMINATOR.

|  | Model | Accuracy [%] | Recall [%] | Precision [%] | F1 score | Hz | Memory [MB] |
|---|---|---|---|---|---|---|---|
| base line method[5] (unsupervised learning) | R | 57.25 | 66.75 | 56.09 | 60.95 | 0.127 | 323 |
|  | D | 59.75 | 58.50 | 60.00 | 59.24 | 0.127 | 323 |
|  | R+D | 60.00 | 72.25 | 58.03 | 64.36 | 0.125 | **323** |
| our method (unsupervised learning) | R | 67.88 | 66.75 | 68.29 | 67.51 | **175.18** | 339 |
|  | D | 72.00 | **77.50** | 69.81 | **73.46** | 102.381 | 352 |
|  | R+D | **72.50** | 68.50 | **74.46** | 71.35 | 93.07 | 354 |
| base line method [38] (supervised learning) | ResNet 50 | 91.63 | 95.75 | 87.64 | 91.52 | 34.46 | 705 |
|  | ResNet 152 | 92.25 | 94.75 | 90.23 | 92.44 | 12.21 | 1357 |
| our method (supervised learning) | R | 85.38 | 83.50 | 86.75 | 85.10 | 175.17 | 338 |
|  | D | 91.63 | 94.50 | 89.36 | 91.86 | 103.17 | 356 |
|  | F | 92.25 | 95.50 | 89.67 | 92.49 | **329.37** | **326** |
|  | R+D | 91.63 | 94.00 | 89.74 | 91.81 | 94.11 | 358 |
|  | D+F | 93.00 | **96.50** | 90.19 | 93.24 | 96.41 | 357 |
|  | R+F | 93.13 | 95.00 | 91.56 | 93.25 | 119.99 | 348 |
|  | **R+D+F** | **94.25** | 95.75 | **92.96** | **94.33** | 89.69 | 359 |

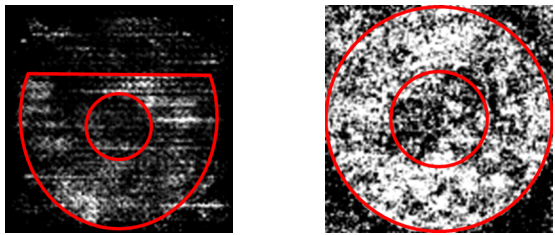

[a] mean of saliency map  [b] weight for Residual loss

Fig. 9. Visualization of proposed neural network.

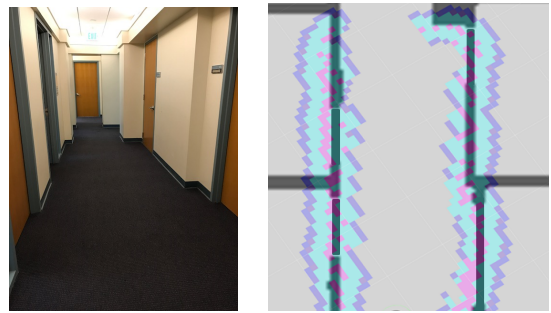

Fig. 10. GoNoGo Cost map generated by our method. Left side photo shows the overview of the environment for the experiment. Right side photo shows the cost map based on our method.

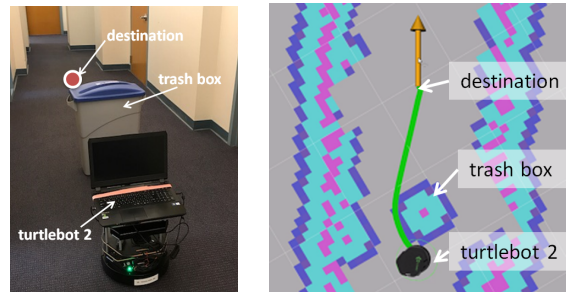

Fig. 11. Planing based on our cost map. Left side photo shows the overview of the environment for the experiment. Right side photo shows the cost map and trajectory generated by the global planner.

*2) Near supervised learning:* We chose pre-trained ResNet 50 and 152 on ImageNet [38] as the supervised baseline method for our near unsupervised learning approach. We extracted features from ResNet and trained the last FC layer to decide GO or NO GO using this features. We only trained the FC layer because training the whole neural network on the small dataset of positive and negative examples would lead to over-fitting.

As can be seen in Table IV, our method (R+D+F) outperforms the baseline supervised methods. In addition, the baseline ResNets need 2 to 4 times more memory and significantly more computation time than our method (R+D+F). Our approach can run at 89.69 Hz, which is much faster than the camera's frame rate of 15 fps.

*G. Cost-map GoNoGo*

One of the practical applications of our approach is building cost-maps for navigation. For example, Figure 10 shows a cost-map that was generated for static objects in the environment of the robot. This cost-map was built by tele-operating the robot and placing high-cost obstacles in its cost-map whenever the view from its camera detected a NO GO situation. These obstacles correspond to the magenta parts of the map in the right image of Fig. 10. The surrounding of the obstacles (considering a fixed radius for inflation) was set to a lower, but still considerable cost for navigation (blue and light blue areas in the map). Note that any automatic exploration method, such as Frontier exploration [39], could have been used as well to generate this cost-map.

With the proposed approach, it is also possible to build cost-maps of dynamic environments. For example, Fig.11 shows the path planned for the robot (green line) based on a cost-map that was generated in real-time using the output of our classifier. In this experiment, we placed a trash box in front of the robot, in between its location and a destination goal. The robot then adjusted its planned trajectory to avoid the obstacle and reach the desired goal. Note that for these experiments, we used the Ricoh THETA camera only and no other sensors.

## V. CONCLUSION

We proposed an unsupervised and a near unsupervised learning approach to classify GO and NO GO scenarios observed from a fish eye RGB camera on a robot. Our approach outperformed baseline methods regarding performance (accuracy, recall, precision, and f1 score) and on computational requirements (calculation speed and memory footprint). We also showed that our method could be used to generate cost-maps for robot navigation.

In terms of future work, more experiments are needed to validate the effectiveness of the proposed approach in other scenarios not considered in the present work, like outdoor environments. We would also like to test our method on other robots and further evaluate obstacle avoidance.

## VI. ACKNOWLEDGEMENTS

We appreciate TOYOTA Central R & D Labs., INC. for the financial support to Noriaki Hirose as visiting scholar in Stanford University. In addition, we appreciate Fei Xia and Kazuki Kozuka for the helpful discussions.

## REFERENCES

[1] Thrun, Sebastian, et al. "MINERVA: A second-generation museum tour-guide robot." Proc. of the IEEE international conference on Robotics and automation, Vol. 3, 1999.
[2] N. Hirose et al., "Personal Robot Assisting Transportation to Support Active Human Life – Posture Stabilization based on Feedback Compensation of Lateral Acceleration–", Proc. of the IEEE/RSJ International Conference on Intelligent Robots & Systems, pp.659–664, 2013.
[3] N. Hirose et al., "Personal Robot Assisting Transportation to Support Active Human Life – Reference Generation based on Model Predictive Control for Robust Quick Turning –", Proc. of IEEE International Conference on Robotics & Automation, pp.2223–2230, 2014.
[4] N. Hirose et al., "Personal Robot Assisting Transportation to Support Active Human Life –Following Control based on Model Predictive Control with Multiple Future Predictions–", Proc. of the IEEE/RSJ International Conference on Intelligent Robots & Systems, pp.5395–5402, 2015.
[5] T. Schlegl et al., "Unsupervised Anomaly Detection with Generative Adversarial Networks to Guide Marker Discovery", Proc. of the International Conference on Information Processing in Medical Imaging, pp.146–157, 2017.
[6] I, J, Goodfellow et al, "Generative Adversarial Networks" Proc. of Advances in neural information processing systems, pp.2672–2680, 2014.
[7] A. Radford et al, "Unsupervised Representation Learning with Deep Convolutional Generative Adversarial Networks", arXiv:1607.07539, 2015.
[8] J. Y. Zhu et al, "Generative Visual Manipulation on the Natural Image Manifold", arXiv:1609.03552, 2016.
[9] B. Suger et al, "Traversability Analysis for Mobile Robots in Outdoor Environments: A Semi- Supervised Learning Approach Based on 3D-Lidar Data", Proc. of IEEE International Conference on Robotics and Automation, pp.3941–3946, 2015.
[10] M. Pfeiffer et al, "From perception to decision: A data-driven approach to end-to-end motion planning for autonomous ground robots", Proc. of IEEE International Conference on Robotics and Automation, pp.1527–1533, 2017.
[11] J. Borenstein et al, "Real-Time Obstacle Avoidance for Fast Mobile Robots", IEEE trans. on Systems, Man, and Cybernetics, Vol. 19, No. 5, pp.1179–1187, 1989.
[12] M. J. Er et al, "Obstacle Avoidance of a Mobile Robot Using Hybrid Learning Approach", IEEE trans on Industrial Electronics, Vol. 52, No. 3, pp.898–905, 2005.
[13] I. Ulrich et al, "Appearance-Based Obstacle Detection with Monocular Color Vision" Proc. of AAAI Conference on Artificial Intelligence, 2000.
[14] H. Alvarez et al, "Collision avoidance for quadrotors with a monocular camera" The 14th International Symposium on Experimental Robotics, pp.195–209, 2015.
[15] K.S. Shankar et al, "Vision and Learning for Deliberative Monocular Cluttered Flight" Results of the 10th International Conference, pp.391-409, 2016.
[16] S. Daftry et al, "Robust Monocular Flight in Cluttered Outdoor Environments", arXiv:1604.04779, 2016.
[17] A. Sadeghian et al, "Tracking The Untrackable: Learning to Track Multiple Cues with Long-Term Dependencies", arXiv: 1701.01909, 2017.
[18] D. N. Tuong et al, "Model learning for robot control: a survey", Cognitive Processing, pp.319–340, 2011.
[19] N. Hirose et al, "Modeling of rolling friction by recurrent neural network using LSTM" Proc. of the International Conference on Robotics and Automation, pp. 6471–6478, 2017.
[20] Robicquet, Alexandre, et al. "Learning social etiquette: Human trajectory understanding in crowded scenes."' European conference on computer vision. Springer International Publishing, 2016.
[21] T. Zhang et al, "Learning Deep Control Policies for Autonomous Aerial Vehicles with MPC-Guided Policy Search", arXiv:1509.06791, 2015.
[22] Y. LeCun, "Off-Road Obstacle Avoidance through End-to-End Learning", Proc. of Neural Information Processing Systems, 2006.
[23] S. Ross et al, "Learning Monocular Reactive UAV Control in Cluttered Natural Environments", Proc. of the International Conference on Robotics and Automation, pp.1765–1772, 2013.
[24] S. Ross et al, "A reduction of imitation learning and structured prediction to no-regret online learning", Proc of the 14th International Conference on Articial Intelligence and Statistics, pp.627–635, 2011.
[25] L. Tai et al, "A Deep-Network Solution Towards Model-less Obstacle Avoidance", Proc of IEEE/RSJ International Conference on Intelligent Robots and Systems, pp.2759–2764.
[26] B. Q. Huang et al, "Reinforcement Learning Neural Network to the Problem of Autonomous Mobile Robot Obstacle Avoidance", Proc. of the 4th International Conference on Machine Learning and Cybernetics, pp.85–89, 2005.
[27] A. Giusti et al, "A Machine Learning Approach to Visual Perception of Forest Trails for Mobile Robots", IEEE Robotics and Automation Letters, Vol.1, No.2, pp.661–667, 2016.t
[28] D. Gandhi et al, "Learning to Fly by Crashing", arXiv: 1704.05588, 2017.
[29] C. Elkan et al, "Learning Classifiers from only positive and unlabeled examples", Proc. of the 9th International Conference on Information Processing and Management of Uncertainty in Knowledge-Based Systems, 2002.
[30] D. A. Clevert et al, "Fast and Accurate Deep Neural Network Learning by Exponential Linear Units(ELUs)", arXiv:1511.07289, 2016.
[31] http://www.turtlebot.com/turtlebot2/
[32] https://theta360.com/en/
[33] Quigley, Morgan, et al. "ROS: an open-source Robot Operating System."' ICRA workshop on open source software. Vol. 3. No. 3.2. 2009.
[34] https://www.google.com/maps
[35] https://chainer.org/
[36] D. P. Kingma et al, "ADAM: A Method for Stochastic Optimization", arXiv: 1412.6980, 2017.
[37] K. Simonyan et al, "Deep Inside Convolutional Networks: Visualizing Image Classification Models and Saliency Maps"', arXiv: 1312.6034, 2014.
[38] K. He et al, "Deep Residual Learning for Image Recognition", arXiv: 1512.03385, 2015.
[39] B. Yamauchi, "A frontier-based approach for autonomous exploration"', Computational Intelligence in Robotics and Automation, 1997. CIRA'97., Proceedings., 1997 IEEE International Symposium on (pp. 146-151).